\documentclass[10pt,twocolumn]{article}
\usepackage[T1]{fontenc}
\usepackage[utf8]{inputenc}
\usepackage[margin=0.75in,columnsep=0.25in]{geometry}
\usepackage{amsmath,amssymb}
\usepackage{graphicx}
\usepackage{booktabs}
\usepackage{float}
\usepackage{microtype}
\usepackage{xcolor}
\usepackage{tikz}
\usetikzlibrary{arrows.meta}
\usepackage{algorithm}
\usepackage{algpseudocode}
\algrenewcommand\algorithmiccomment[1]{\hfill{\color{dblue}$\triangleright$ #1}}
\definecolor{dblue}{RGB}{38,84,200}
\definecolor{dgray}{RGB}{200,197,190}
\definecolor{dred}{RGB}{238,108,122}
\definecolor{dsky}{RGB}{170,198,245}
\definecolor{dlav}{RGB}{226,214,246}
\definecolor{dpeach}{RGB}{250,222,176}
\usepackage[numbers,sort&compress]{natbib}
\usepackage[font=small,labelfont=bf]{caption}
\usepackage[colorlinks=true,linkcolor=black,citecolor=blue!50!black,urlcolor=blue!50!black]{hyperref}

\newcommand{\pct}{\,\%}
\newcommand{\scratchseeds}{one seed}

\title{\vspace{-1.5em}A Link Budget for Learning\\What Multi Datacenter Training Needs from Long Haul Fiber}
\author{Pranav Reddy\\University of California, Santa Barbara\\\texttt{preddy@ucsb.edu}}
\date{}

\begin{document}
\maketitle

\begin{abstract}
Distributed training methods such as DiLoCo train one model across several
datacenters, and each synchronization sends a model sized update over long haul
fiber. Optical links are engineered to deliver error free bits at 99.999\pct{}
availability. We measure what training needs
from the link by training through a simulated coherent link, and we state the need
in the units of a link budget. On a small task trained to convergence over 8 seeds,
the update needs 17\,dB of fidelity, which is 5 bits per value, and the link may be
down for 2\pct{} of rounds. At 1000\,km, sending 5 bits in place of 16 raises the
update rate per wavelength by a factor of 3.0, and relaxing availability adds a
further 6\pct. The Shannon bound for the same need is 2.8 times higher still.
Unequal error protection carries fewer updates than a shorter code on a standard
link. In one run of a 70M parameter language model, early training tolerates a
1-bit code, while a bit error rate of $10^{-6}$ on the scale bits ends the run. All
results come from simulation.
\end{abstract}

\section{Introduction}

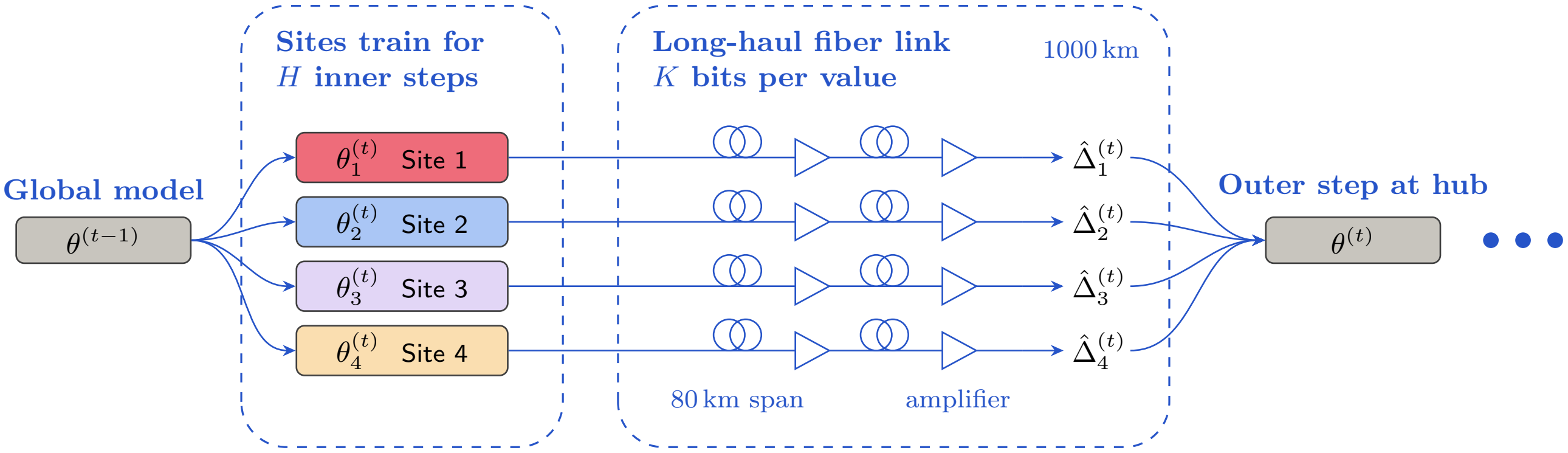
\begin{figure*}[t]
\centering
\begin{tikzpicture}[x=1cm,y=1cm,font=\small,>={Stealth[length=4pt,width=3.5pt]},
  mbox/.style={draw=black!75,line width=0.5pt,rounded corners=2.5pt,minimum height=0.5cm,inner sep=2pt},
  grp/.style={draw=dblue,dashed,line width=0.6pt,rounded corners=14pt},
  lab/.style={text=dblue,font=\small\bfseries,align=left},
  flow/.style={draw=dblue,line width=0.5pt,->},
  coil/.pic={\draw[dblue,line width=0.5pt,fill=white] (-0.09,0.17) circle (0.17);
             \draw[dblue,line width=0.5pt] (0.09,0.17) circle (0.17);},
  amp/.pic={\draw[dblue,line width=0.5pt,fill=white] (-0.17,-0.2)--(0.2,0)--(-0.17,0.2)--cycle;}]
  \node[lab,anchor=south] at (1.0,0.32) {Global model};
  \node[mbox,fill=dgray,minimum width=1.9cm] (g0) at (1.0,0) {$\theta^{(t-1)}$};
  \draw[grp] (2.5,-2.25) rectangle (6.0,2.55);
  \node[lab,anchor=north west] at (2.75,2.4) {Sites train for\\$H$ inner steps};
  \foreach \y/\i/\c in {0.9/1/dred,0.2/2/dsky,-0.5/3/dlav,-1.2/4/dpeach} {
    \node[mbox,fill=\c,minimum width=2.3cm] (s\i) at (4.25,\y) {$\theta_{\i}^{(t)}$\hspace{0.7em}\textsf{\footnotesize Site \i}};
    \draw[flow] (g0.east) to[out=0,in=180] (s\i.west);
  }
  \draw[grp] (6.6,-2.25) rectangle (12.6,2.55);
  \node[lab,anchor=north west] at (6.85,2.4) {Long-haul fiber link\\$K$ bits per value};
  \node[text=dblue,font=\footnotesize,anchor=north east] at (12.4,2.3) {1000\,km};
  \foreach \y/\i in {0.9/1,0.2/2,-0.5/3,-1.2/4} {
    \draw[dblue,line width=0.5pt] (s\i.east)--(11.0,\y);
    \draw[flow] (11.0,\y)--(11.45,\y);
    \foreach \x in {7.9,9.5} {\pic at (\x,\y) {coil};}
    \foreach \x in {8.7,10.3} {\pic at (\x,\y) {amp};}
    \node[anchor=west,inner sep=1pt] (d\i) at (11.5,\y) {$\hat\Delta_{\i}^{(t)}$};
  }
  \node[text=dblue,font=\footnotesize] at (7.9,-1.75) {80\,km span};
  \node[text=dblue,font=\footnotesize] at (10.3,-1.75) {amplifier};
  \node[lab,anchor=south] at (14.6,0.32) {Outer step at hub};
  \node[mbox,fill=dgray,minimum width=1.9cm] (g1) at (14.6,0) {$\theta^{(t)}$};
  \foreach \i in {1,2,3,4} {\draw[flow] (d\i.east) to[out=0,in=180] (g1.west);}
  \foreach \x in {16.1,16.45,16.8} {\fill[dblue] (\x,0) circle (0.085);}
\end{tikzpicture}
\caption{One round of training through the link. The global model
$\theta^{(t-1)}$ is copied to $k$ sites (here $k{=}4$), and each site trains its
replica $\theta_i^{(t)}$ for $H$ steps. Each site sends its update over a simulated
long haul fiber link at $K$ bits per value, and the hub receives a distorted copy
$\hat\Delta_i^{(t)}$. The hub averages the copies and takes the outer step to
produce $\theta^{(t)}$, which returns to the sites over the same link.}
\label{fig:system}
\end{figure*}

Large models are normally trained in one datacenter, because every optimizer step
exchanges gradients among all accelerators. DiLoCo~\cite{douillard2023diloco}
relaxes this by letting each site train on its own for $H$ steps and then exchange
one update (Figure~\ref{fig:system}). Streaming DiLoCo~\cite{douillard2025streaming} staggers
the exchange and quantizes it to 4 bits per value. In both methods the update
travels between sites over a long haul optical link.

Long haul links are designed with a link budget. The budget starts from the
signal-to-noise ratio (SNR) the fiber can deliver and subtracts impairments and
margin until the link is error-free for 99.999\pct{} of the time. Training has a
different tolerance. It already averages noisy gradients, and DiLoCo is known to
survive quantized updates and dropped
workers~\cite{douillard2023diloco,douillard2025streaming}. We therefore ask what a
training run needs from the link, in decibels and in availability, and what a link
built to that need would carry. Our findings are as follows.

\begin{itemize}\itemsep0pt
\item \textbf{Most of the gain comes from sending fewer bits.} On our converged
task, 5 bits per value carry 3.0 times as many update values as 16. Streaming DiLoCo
already sends 4 bits, so this is the baseline for any link level proposal.
\item \textbf{Unequal error protection carries less than a shorter code.} A bit
that can tolerate a 10\pct{} error rate adds less than it costs to send.
\item \textbf{Only the scale bits need exact delivery.} Errors in the remaining
bits behave like added noise of the same power.
\item \textbf{Early language model training tolerates about 1 bit per value}, and
it fails in the same way when scale bits are corrupted.
\end{itemize}

\section{A Link Budget for Learning}

\paragraph{Setup.} We simulate $K{=}4$ sites and a hub. Each site sends its
pseudo gradient, which is the change in its weights over $H$ local steps. The hub
averages the four, takes the outer optimizer step, and returns the result. Both
directions pass through the simulated link. Algorithm~\ref{alg:link} gives the
round in full, including the block code $Q_K$ and the two measures against replica
drift that we test in Section~\ref{sec:exp}.

\begin{algorithm}[!tb]
\caption{DiLoCo through a $K$ bit link with a tracked downlink}
\label{alg:link}
\begin{algorithmic}[1]
\Require Global model $\theta^{(0)}$, $k$ sites with replicas $\phi_i \gets \theta^{(0)}$
\Require Optimizers \textsc{InnerOpt} and \textsc{OuterOpt}
\Require Block code $Q_K$ with $K$ bits per value, resync period $R$
\For{round $t = 1 \ldots T$}
  \For{site $i = 1 \ldots k$ whose link is up}
    \State $\theta_i^{(t)} \gets$ $H$ steps of \textsc{InnerOpt} from $\phi_i$
    \State $\hat\Delta_i^{(t)} \gets \textsc{Link}\big(Q_K(\phi_i - \theta_i^{(t)})\big)$ \Comment{uplink}
  \EndFor
  \State $\Delta^{(t)} \gets$ mean of the $\hat\Delta_i^{(t)}$ that arrived
  \State $u \gets \textsc{OuterOpt}(\Delta^{(t)})$
  \State $\hat u \gets Q_K(u)$ \Comment{hub decodes its own update}
  \State $\theta^{(t)} \gets \theta^{(t-1)} + \hat u$
  \State $\phi_i \gets \phi_i + \hat u$ for every site that is up \Comment{downlink}
  \If{$t \bmod R = 0$}
    \State $\phi_i \gets \theta^{(t)}$ sent exactly \Comment{resync}
  \EndIf
\EndFor
\Statex
\Function{$Q_K$}{$x$} \Comment{encode, then decode}
  \For{each block $b$ of 32 values of $x$}
    \State $s \gets$ RMS of $b$, rounded to 8 bits \Comment{exact}
    \State $q_j \gets$ sign of $b_j$ and $K{-}1$ bits of $|b_j|/s$ \Comment{noisy}
  \EndFor
  \State \Return the values rebuilt from every $s$ and $q_j$
\EndFunction
\end{algorithmic}
\end{algorithm}

\begin{figure}[t]
\centering
\begin{tikzpicture}[x=0.44cm,y=0.45cm,font=\footnotesize,
  cell/.style={draw=black!75,line width=0.4pt,rounded corners=1pt},
  exact/.style={cell,fill=dred},noisy/.style={cell,fill=dsky}]
  \node[anchor=west,font=\small\bfseries,text=dblue] at (0,3.1) {bf16};
  \node[anchor=west] at (2.4,3.1) {16 bits per value};
  \draw[noisy] (0,1.6) rectangle (1,2.5);
  \foreach \i in {1,...,8} {\draw[exact] (\i,1.6) rectangle (\i+1,2.5);}
  \foreach \i in {9,...,15} {\draw[noisy] (\i,1.6) rectangle (\i+1,2.5);}
  \node[below] at (0.5,1.55) {sign}; \node[below] at (5,1.55) {exponent}; \node[below] at (12.5,1.55) {mantissa};
  \node[anchor=west,font=\small\bfseries,text=dblue] at (0,-0.6) {Block code};
  \node[anchor=west] at (5.2,-0.6) {$K$ bits per value, one scale per 32 values};
  \foreach \i in {0,...,3} {\draw[exact] (\i,-2.1) rectangle (\i+1,-1.2);}
  \foreach \g in {0,1,2} {\foreach \i in {0,1,2} {\draw[noisy] (4.5+\g*3.4+\i,-2.1) rectangle (5.5+\g*3.4+\i,-1.2);}}
  \node at (15.2,-1.65) {$\cdots$};
  \node[below] at (2,-2.15) {shared scale}; \node[below] at (6,-2.15) {value 1}; \node[below] at (9.4,-2.15) {value 2};
  \node[below] at (12.8,-2.15) {value 3};
  \draw[exact] (0,-4.6) rectangle (0.7,-3.95); \node[anchor=west] at (0.9,-4.28) {must arrive exactly};
  \draw[noisy] (8.2,-4.6) rectangle (8.9,-3.95); \node[anchor=west] at (9.1,-4.28) {may be noisy};
\end{tikzpicture}
\caption{The two update formats we compare. The block code follows microscaling
formats~\cite{rouhani2023mx} and shares one scale among 32 values. We write
``mx8'' for the block code with $K{=}8$.}
\label{fig:bits}
\end{figure}
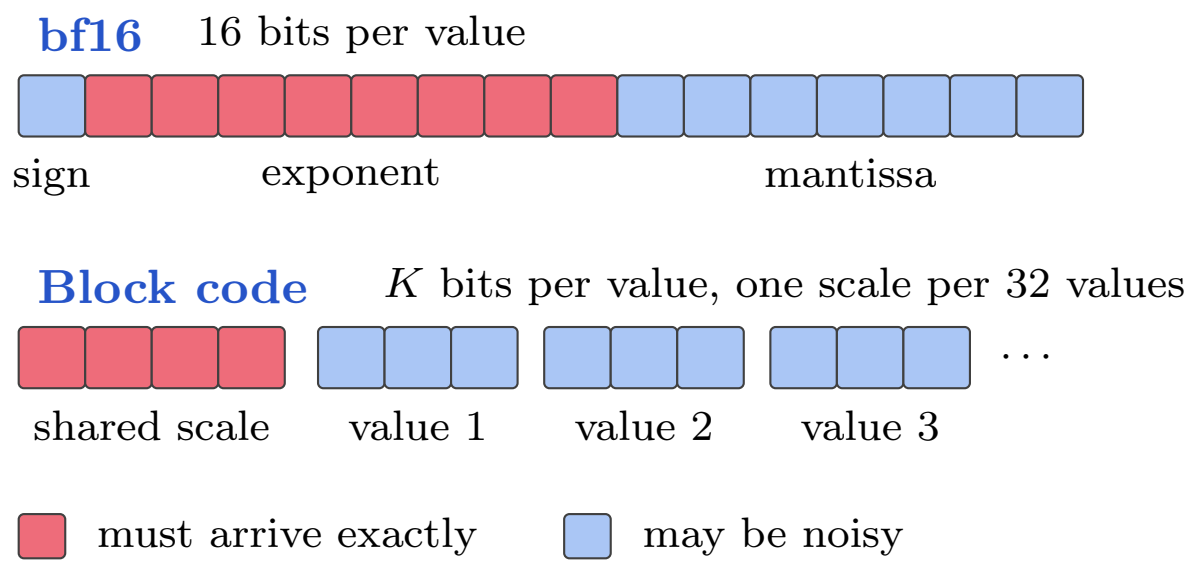

\paragraph{Three needs.} We describe what training needs from the link with three
quantities. \emph{Fidelity} is the signal-to-distortion ratio of a received update,
$\mathrm{SDR}=\lVert u\rVert^2/\lVert\hat u-u\rVert^2$, where $u$ is the update
sent and $\hat u$ the update received. It measures quantization, bit errors and
added noise on one scale, and it determines the bits per value.
\emph{Availability} is the share of rounds in which a site's update arrives. It
determines the SNR margin (Figure~\ref{fig:budget}). \emph{Integrity} is the exact
delivery of the few bits whose errors are unbounded (Figure~\ref{fig:bits}).

\begin{figure}[t]
\centering
\begin{tikzpicture}[x=0.63cm,y=0.56cm,font=\footnotesize]
  \def\wbar#1#2#3#4#5{
    \draw[draw=black!75,line width=0.4pt,rounded corners=1.5pt,fill=#5] (#2-12,-#1-0.3) rectangle (#3-12,-#1+0.3);
    \node[anchor=east] at (-0.15,-#1) {#4};}
  \foreach \x in {12,14,16,18,20} {\draw[gray!40,line width=0.3pt] (\x-12,0.7)--(\x-12,-6.9);
     \node[below,gray!50!black] at (\x-12,-6.9) {\x};}
  \node[below,gray!50!black] at (4,-7.8) {SNR (dB)};
  \wbar{0}{12}{19.35}{Amplifier noise only}{dgray}
  \wbar{1}{17.58}{19.35}{Nonlinearity}{dlav}
  \wbar{2}{15.95}{17.58}{Transceiver}{dlav}
  \wbar{3}{15.46}{15.95}{Ageing}{dlav}
  \wbar{5}{13.33}{15.46}{Margin, 99.999\,\%}{dpeach}
  \wbar{6}{14.43}{15.46}{Margin, 98\,\%}{dsky}
  \node[anchor=west] at (7.4,0) {19.3};
  \node[anchor=west] at (7.4,-1) {$-1.8$}; \node[anchor=west] at (5.63,-2) {$-1.6$}; \node[anchor=west] at (4.0,-3) {$-0.5$};
  \draw[dblue,dashed,line width=0.6pt] (3.46,0.7)--(3.46,-6.6);
  \node[anchor=west,text=dblue] at (3.75,-4) {\textbf{15.5\,dB} at end of life};
  \node[anchor=east] at (1.3,-5) {$-2.1$}; \node[anchor=east] at (2.4,-6) {$-1.0$};
  \node[anchor=west] at (3.6,-5) {conventional};
  \node[anchor=west,text=dblue] at (3.6,-6) {\textbf{for learning}};
\end{tikzpicture}
\caption{Link budget for 1000\,km of standard fiber. Impairments reduce the SNR to
15.5\,dB at end of life. A conventional link then holds 2.1\,dB of margin for
99.999\pct{} availability. Training tolerates 2\pct{} downtime and needs 1.0\,dB.
The margins assume an SNR drift with a standard deviation of 0.5\,dB.}
\label{fig:budget}
\end{figure}
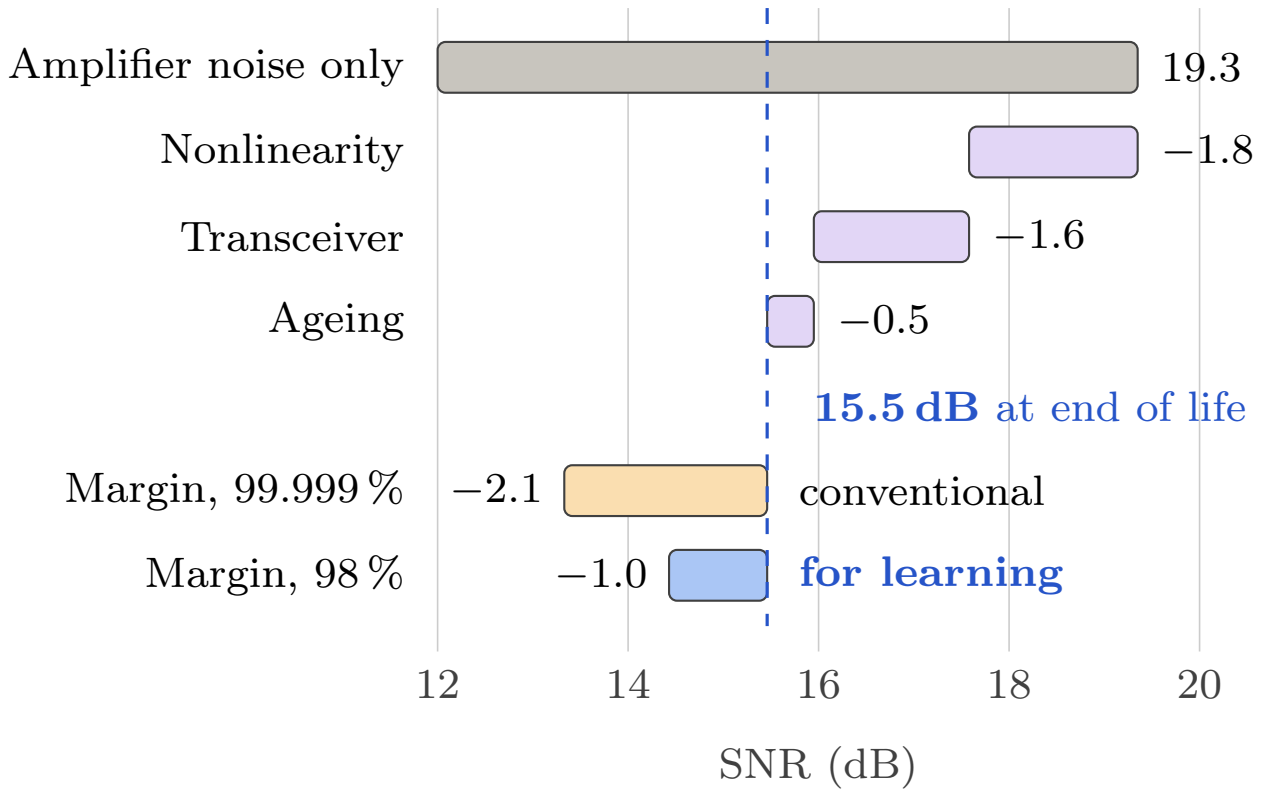

\paragraph{Link model.} The line is a chain of 80\,km spans of G.652 fiber and
amplifiers with a 5\,dB noise figure. Nonlinear interference is computed with the
closed-form GN model~\cite{poggiolini2014gn} at the optimum launch power. The
transceiver runs at 130\,GBd and its own noise limits the SNR to 21\,dB. Forward
error correction (FEC) is modeled as a threshold with 25\pct{} overhead and a
pre FEC bit error rate (BER) of $2\times10^{-2}$. The SNR drifts slowly and
occasionally drops out, and residual errors arrive either scattered or in bursts.
The FEC and drift parameters are assumptions and are not fitted to a deployed
system.

\paragraph{Yardstick.} A symbol at SNR $\gamma$ carries at most
$2\log_2(1+\gamma)$ bits. A Gaussian source needs $\tfrac12\log_2\mathrm{SDR}$
bits per value to reach a given fidelity~\cite{shannon1959fidelity}. The ratio of
the two is an upper bound on update values per symbol, and we report every
transport against it.

\section{Experiments}
\label{sec:exp}

We report the \emph{penalty}, which is the relative increase in final validation
loss over an ideal link run with the same seed. We use three settings. The
\emph{toy task} is a 22k-parameter MLP on a non-IID teacher student problem with
$H{=}50$ and a Nesterov outer step. It is trained to convergence over 8 seeds. The
\emph{from scratch} setting trains Pythia-70M~\cite{biderman2023pythia} from random
weights on WikiText-2~\cite{merity2017wikitext} with $H{=}8$, using \scratchseeds.
The \emph{fine-tuning} setting starts from pretrained Pythia-160M, 410M and 1B and
runs 30 rounds with $H{=}8$ and plain averaging.

\begin{figure*}[t]
\centering
\includegraphics[width=\textwidth]{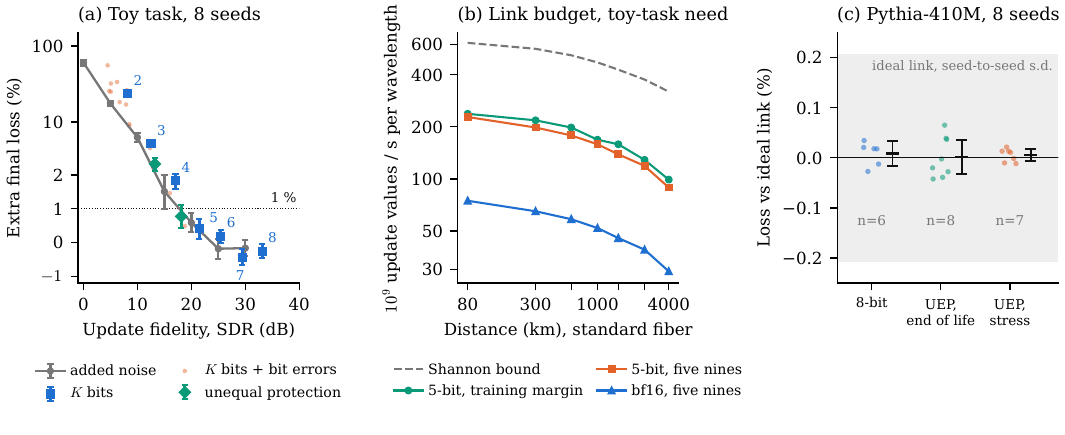}
\caption{(a) Toy task. Penalty against the fidelity the update arrived with. Error
bars are standard errors over 8 seeds and the numbers beside the squares give $K$.
(b) Update rate per wavelength against distance for the toy task need. (c)
Pythia-410M fine tuning. Each dot is one seed compared with its own ideal link run,
and the bars show the mean and 95\pct{} interval. The band is the seed to seed
standard deviation of the ideal link, 0.21\pct.}
\label{fig:results}
\end{figure*}

\paragraph{How many bits per value?} With added Gaussian noise, the penalty
crosses 1\pct{} near 17\,dB of SDR (Figure~\ref{fig:results}a). A 5-bit block code
delivers 21\,dB and costs $+0.4\pm0.3$\pct. A 4-bit code delivers 17\,dB and costs
$+1.8\pm0.3$\pct. We take 5 bits as the need of this task.

\paragraph{Does unequal error protection help?} Bit errors in the element bits
fall on the same curve as added noise, so SDR predicts their cost. We tested an
8-bit format whose low bits travel on lightly protected lanes. It delivers 18\,dB,
which is the quality of a 4-bit or 5-bit code, while sending 8.25 bits per value
(Table~\ref{tab:toy}). The 5-bit code on a standard lane carries 31\pct{} more
values at a penalty that is equal or lower.

\begin{table}[t]
\centering\small
\caption{Toy task after 40 rounds, 8 seeds, mean $\pm$ standard error. The link is
1000\,km at 12.4\,dB SNR. Rate is in $10^9$ update values per second per
wavelength.}
\label{tab:toy}
\begin{tabular}{@{}lrrr@{}}
\toprule
Transport & SDR (dB) & Penalty (\%) & Rate \\
\midrule
8-bit, error-free & 41.3 & $-0.3\pm0.2$ & 95 \\
8-bit, unequal protection & 18.1 & $+0.8\pm0.4$ & 114 \\
5-bit, error-free & 21.4 & $+0.4\pm0.3$ & 149 \\
4-bit, error-free & 17.1 & $+1.8\pm0.3$ & 184 \\
3-bit, error-free & 12.5 & $+5.2\pm0.4$ & 240 \\
\bottomrule
\end{tabular}
\end{table}

\paragraph{Which bits must arrive exactly?} The block scales are the exception to
the noise rule. At a scale BER of $10^{-3}$, all 8 toy runs diverge when the errors
are scattered, and none diverge when the same number of errors arrives in 16\,kb
bursts. The damage therefore depends on the number of error events and is not
predicted by the BER. The requirement on the link is an errored-frame rate, and a
frame that fails its check must be dropped (Appendix~\ref{app:detail}).

\paragraph{Do errors accumulate?} Each site applies its own distorted copy of the
update, so the replicas drift apart. With unequal protection the penalty grows from
$+0.8\pm0.4$\pct{} at 40 rounds to $+28.6\pm3.7$\pct{} at 640. For quantization
there is a fix that costs nothing. The hub applies the update as the sites decode
it, and the 4-bit and 5-bit codes then show no penalty at 640 rounds. Random bit
errors differ at each site and still need a periodic exact resynchronization.

\paragraph{How much downtime is tolerable?} Missing 2\pct{} of rounds costs
$+0.8\pm1.3$\pct, and missing 10\pct{} costs $+2.4\pm1.1$\pct. We take 2\pct{}
as the availability need.

\paragraph{What does the budget buy?} Table~\ref{tab:budget} converts these needs
into update rate at 1000\,km, and Figure~\ref{fig:results}b shows other distances.
Moving from 16 to 5 bits is worth a factor of 3.0. Matching the margin to training
adds 6\pct{} at 1000\,km and between 4 and 14\pct{} from 80 to 4000\,km. The
remaining factor of 2.8 to the Shannon bound has two parts. Source coding accounts
for a factor of 1.8, because we send 5.25 bits where rate-distortion theory needs
2.9. Channel coding accounts for a factor of 1.5, because the modem carries 6.8
bits per symbol where capacity is 10.4.

\begin{table}[t]
\centering\small
\caption{Update rate per wavelength at 1000\,km for the toy-task need. Each step
is relative to the row above.}
\label{tab:budget}
\begin{tabular}{@{}lrl@{}}
\toprule
Transport & $10^9$ values/s & Step \\
\midrule
bf16, error-free, five-nines margin & 52 & reference \\
5-bit, error-free, five-nines margin & 158 & $\times3.0$ \\
5-bit, margin matched to training & 168 & $+6$\pct \\
Shannon bound at 17\,dB SDR & 471 & $\times2.8$ \\
\bottomrule
\end{tabular}
\end{table}

\paragraph{Does a language model need the same?} Table~\ref{tab:lm} repeats the
measurements on Pythia-70M trained from scratch. We ran full runs for rounds 0 to
50, and 40-round continuations from ideal-link checkpoints at rounds 100 and 300.
Noise as strong as the update itself (0\,dB SDR) and a 1-bit code both stay within
run-to-run variation in every window. The need does not tighten over 340 rounds,
during which the loss falls from 10.1 to 5.7. Element bits tolerate a BER of
$10^{-2}$. A BER of $10^{-6}$ on the block scales ends the run.

The coarse codes score slightly below the ideal link, and this is a step-size
effect. The 1-bit code shrinks each update by a factor of 0.64, and an ideal link
whose outer step is scaled by 0.64 improves by the same amount. A compressed link
should therefore be compared with a baseline whose outer step has been tuned.

\begin{table}[t]
\centering\small
\caption{Pythia-70M from scratch. Loss relative to the ideal link in percent,
averaged over the last 5 rounds of each window, with a 95\pct{} interval where
$n>1$. Quantized codes use the tracked downlink. Short and long outages last about
1 and 8 rounds.}
\label{tab:lm}
\begin{tabular}{@{}lrrr@{}}
\toprule
Link & 0--50 & 100--140 & 300--340 \\
 & ($n=1$) & ($n=1$) & ($n=1$) \\
\midrule
Noise, 10\,dB SDR & $+0.1$ & ${-}0.1$ & ${-}0.3$ \\
Noise, 0\,dB SDR & $+0.4$ & ${-}0.0$ & ${-}0.2$ \\
Noise, $-10$\,dB SDR & $+2.4$ &  &  \\
4-bit block code &  & ${-}0.2$ & ${-}0.2$ \\
2-bit block code &  & ${-}0.6$ & ${-}0.5$ \\
1-bit block code & ${-}0.9$ & ${-}0.9$ & ${-}1.3$ \\
Sign and exponent only & $+0.5$ &  &  \\
\midrule
5\pct{} of rounds missed & $+0.8$ &  &  \\
20\pct{} missed, short & $+2.0$ &  &  \\
20\pct{} missed, long & $+2.2$ &  &  \\
\midrule
Element bits, BER $10^{-2}$ & ${-}0.2$ &  &  \\
Block scales, BER $10^{-6}$ & 1/1 diverged &  &  \\
\midrule
Control, new data order &  & ${-}0.2$ & ${-}0.0$ \\
Control, step $\times0.64$ &  &  & ${-}1.4$ \\
\bottomrule
\end{tabular}

\end{table}

These runs cover early training only. The model is far from converged, and the
same 1-bit code fails on the converged toy task. Taken together, our experiments
support a need between 1 and 5 bits per value and do not locate it more closely.

\paragraph{Does the result hold at larger sizes?} Table~\ref{tab:ft} fine-tunes
larger models through the unequal-protection link. No mean gap exceeds 0.1\pct. The
16-bit format with every bit exposed to a BER of $10^{-6}$ diverged in the first
round in all 10 runs. These fine-tuning runs are short and start near a minimum, so
they confirm the scale-bit failure at size and cannot rank link designs. At 1B all
four unequal-protection gaps are positive, and two seeds are too few to tell a
small cost from chance.

\begin{table}[t]
\centering\small
\caption{Fine-tuning for 30 rounds, paired by seed. The gap is the mean with
standard deviation in parentheses (160M), the mean $\pm$ 95\pct{} interval (410M),
or the two per-seed values (1B). UEP is unequal error protection, at end-of-life
BERs ($2.3\times10^{-2}$ on the low bits) or at stress BERs ($10^{-1}$).}
\label{tab:ft}
\begin{tabular}{@{}llrl@{}}
\toprule
Model & Link & Seeds & Gap (\%) \\
\midrule
160M & UEP, end of life & 3 & $+0.05$ (0.11) \\
 & UEP, stress & 3 & $+0.03$ (0.08) \\
\midrule
410M & 8-bit, error-free & 6 & $+0.009\pm0.024$ \\
 & UEP, end of life & 8 & $+0.001\pm0.034$ \\
 & UEP, stress & 7 & $+0.005\pm0.012$ \\
\midrule
1B & 8-bit, error-free & 2 & $+0.00$, $-0.02$ \\
 & UEP, end of life & 2 & $+0.02$, $+0.08$ \\
 & UEP, stress & 2 & $+0.09$, $+0.02$ \\
\bottomrule
\end{tabular}
\end{table}

\section{Related Work}

DiLoCo~\cite{douillard2023diloco} builds on local SGD and federated
averaging~\cite{stich2019local,mcmahan2017fedavg}. Streaming
DiLoCo~\cite{douillard2025streaming} sends 4-bit outer gradients with no mantissa,
and \cite{charles2025scaling} studies how DiLoCo scales with model size. Work on
gradient compression~\cite{alistarh2017qsgd,bernstein2018signsgd,karimireddy2019ef,
tang2021onebitadam,vogels2019powersgd} shows that per-step gradients survive
coarse coding. We did not run these methods as baselines. Unequal error protection
goes back to~\cite{masnick1967uep}. Inside a datacenter,
\cite{abolfathimomtaz2025linkerror} finds that a link BER below $10^{-4}$ is
harmless for per-step exchange. We study infrequent updates over long-haul links.
Analog transport of gradients has been studied for wireless
channels~\cite{amiri2020overtheair}. Our contribution is a single budget that
expresses these choices in link terms.

\section{Limitations}
\begin{itemize}\itemsep0pt
\item All results come from simulation. The drift, burst and FEC parameters are
assumptions, and the line model has not been validated against a planning tool.
\item The 17\,dB and 98\pct{} needs were measured on a 22k-parameter task.
\item The language-model evidence is \scratchseeds{} over 340 rounds, far from
convergence, with an outer step that was not tuned. The fine tuning runs cannot
rank link designs.
\item We study a hub-and-spoke topology and count time in rounds. The cost of
resynchronization is not charged to the link.
\end{itemize}

\section{Conclusion}
A training run needs a fidelity from a long haul link, which sets the bits per
value, and an availability, which sets the margin. Beyond these it needs exact
delivery of the scale bits only. Most of the available rate
comes from sending fewer bits, which quantized DiLoCo already does. A factor of 2.8
remains between a uniform code on a standard modem and the Shannon bound. The
measurement still missing is the fidelity that a language model needs late in
training.

\paragraph{Code and data.} \url{https://github.com/pkarakala/fiber-xdc}

\paragraph{Acknowledgments.} Computing was provided by the Center for Scientific
Computing at UC Santa Barbara and by Kaggle's free GPU notebooks. An AI assistant
was used to develop and write the code in the repository.

{\footnotesize\setlength{\bibsep}{2pt}
\bibliographystyle{unsrtnat}
\bibliography{refs}}

@article{douillard2023diloco,
  author  = {Arthur Douillard and Qixuan Feng and Andrei A. Rusu and Rachita Chhaparia and Yani Donchev and Adhiguna Kuncoro and Marc'Aurelio Ranzato and Arthur Szlam and Jiajun Shen},
  title   = {{DiLoCo}: Distributed Low-Communication Training of Language Models},
  journal = {arXiv preprint arXiv:2311.08105},
  year    = {2023}
}

@article{douillard2025streaming,
  author  = {Arthur Douillard and Yanislav Donchev and Keith Rush and Satyen Kale and Zachary Charles and Zachary Garrett and Gabriel Teston and Dave Lacey and Ross McIlroy and Jiajun Shen and Alexandre Ram{\'e} and Arthur Szlam and Marc'Aurelio Ranzato and Paul Barham},
  title   = {Streaming {DiLoCo} with Overlapping Communication: Towards a Distributed Free Lunch},
  journal = {arXiv preprint arXiv:2501.18512},
  year    = {2025}
}

@inproceedings{charles2025scaling,
  author    = {Zachary Charles and Gabriel Teston and Lucio Dery and Keith Rush and Nova Fallen and Zachary Garrett and Arthur Szlam and Arthur Douillard},
  title     = {Communication-Efficient Language Model Training Scales Reliably and Robustly: Scaling Laws for {DiLoCo}},
  booktitle = {Advances in Neural Information Processing Systems (NeurIPS)},
  year      = {2025},
  note      = {arXiv:2503.09799}
}

@inproceedings{mcmahan2017fedavg,
  author    = {H. Brendan McMahan and Eider Moore and Daniel Ramage and Seth Hampson and Blaise {Ag{\"u}era y Arcas}},
  title     = {Communication-Efficient Learning of Deep Networks from Decentralized Data},
  booktitle = {Proceedings of the 20th International Conference on Artificial Intelligence and Statistics (AISTATS)},
  series    = {PMLR},
  volume    = {54},
  year      = {2017}
}

@inproceedings{stich2019local,
  author    = {Sebastian U. Stich},
  title     = {Local {SGD} Converges Fast and Communicates Little},
  booktitle = {International Conference on Learning Representations (ICLR)},
  year      = {2019}
}

@inproceedings{bernstein2018signsgd,
  author    = {Jeremy Bernstein and Yu-Xiang Wang and Kamyar Azizzadenesheli and Animashree Anandkumar},
  title     = {{signSGD}: Compressed Optimisation for Non-Convex Problems},
  booktitle = {Proceedings of the 35th International Conference on Machine Learning (ICML)},
  series    = {PMLR},
  volume    = {80},
  year      = {2018}
}

@inproceedings{alistarh2017qsgd,
  author    = {Dan Alistarh and Demjan Grubic and Jerry Li and Ryota Tomioka and Milan Vojnovic},
  title     = {{QSGD}: Communication-Efficient {SGD} via Gradient Quantization and Encoding},
  booktitle = {Advances in Neural Information Processing Systems (NeurIPS)},
  year      = {2017}
}

@inproceedings{tang2021onebitadam,
  author    = {Hanlin Tang and Shaoduo Gan and Ammar Ahmad Awan and Samyam Rajbhandari and Conglong Li and Xiangru Lian and Ji Liu and Ce Zhang and Yuxiong He},
  title     = {1-bit {Adam}: Communication Efficient Large-Scale Training with {Adam}'s Convergence Speed},
  booktitle = {Proceedings of the 38th International Conference on Machine Learning (ICML)},
  series    = {PMLR},
  volume    = {139},
  year      = {2021}
}

@inproceedings{vogels2019powersgd,
  author    = {Thijs Vogels and Sai Praneeth Karimireddy and Martin Jaggi},
  title     = {{PowerSGD}: Practical Low-Rank Gradient Compression for Distributed Optimization},
  booktitle = {Advances in Neural Information Processing Systems (NeurIPS)},
  year      = {2019}
}

@inproceedings{karimireddy2019ef,
  author    = {Sai Praneeth Karimireddy and Quentin Rebjock and Sebastian U. Stich and Martin Jaggi},
  title     = {Error Feedback Fixes {SignSGD} and other Gradient Compression Schemes},
  booktitle = {Proceedings of the 36th International Conference on Machine Learning (ICML)},
  series    = {PMLR},
  volume    = {97},
  year      = {2019}
}

@article{rouhani2023mx,
  author  = {Bita Darvish Rouhani and Ritchie Zhao and Ankit More and Mathew Hall and Alireza Khodamoradi and Summer Deng and Dhruv Choudhary and Marius Cornea and Eric Dellinger and Kristof Denolf and others},
  title   = {Microscaling Data Formats for Deep Learning},
  journal = {arXiv preprint arXiv:2310.10537},
  year    = {2023}
}

@inproceedings{biderman2023pythia,
  author    = {Stella Biderman and Hailey Schoelkopf and Quentin Anthony and Herbie Bradley and Kyle O'Brien and Eric Hallahan and Mohammad Aflah Khan and Shivanshu Purohit and USVSN Sai Prashanth and Edward Raff and Aviya Skowron and Lintang Sutawika and Oskar van der Wal},
  title     = {Pythia: A Suite for Analyzing Large Language Models Across Training and Scaling},
  booktitle = {Proceedings of the 40th International Conference on Machine Learning (ICML)},
  series    = {PMLR},
  volume    = {202},
  year      = {2023}
}

@inproceedings{merity2017wikitext,
  author    = {Stephen Merity and Caiming Xiong and James Bradbury and Richard Socher},
  title     = {Pointer Sentinel Mixture Models},
  booktitle = {International Conference on Learning Representations (ICLR)},
  year      = {2017}
}

@article{poggiolini2014gn,
  author  = {Pierluigi Poggiolini and Gabriella Bosco and Andrea Carena and Vittorio Curri and Yanchao Jiang and Fabrizio Forghieri},
  title   = {The {GN}-Model of Fiber Non-Linear Propagation and its Applications},
  journal = {Journal of Lightwave Technology},
  volume  = {32},
  number  = {4},
  pages   = {694--721},
  year    = {2014}
}

@article{abolfathimomtaz2025linkerror,
  author  = {Abbas Abolfathimomtaz and Hamid Ebrahimzad and Chuandong Li},
  title   = {Impact of Optical Communication Link Error on Large-Scale {AI} Training in Data Centers},
  journal = {IEEE Open Journal of the Communications Society},
  volume  = {6},
  pages   = {9748--9763},
  year    = {2025}
}

@article{amiri2020overtheair,
  author  = {Mohammad Mohammadi Amiri and Deniz G{\"u}nd{\"u}z},
  title   = {Machine Learning at the Wireless Edge: Distributed Stochastic Gradient Descent Over-the-Air},
  journal = {IEEE Transactions on Signal Processing},
  volume  = {68},
  pages   = {2155--2169},
  year    = {2020}
}

@article{masnick1967uep,
  author  = {Burt Masnick and Jack Wolf},
  title   = {On Linear Unequal Error Protection Codes},
  journal = {IEEE Transactions on Information Theory},
  volume  = {13},
  number  = {4},
  pages   = {600--607},
  year    = {1967}
}

@article{shannon1959fidelity,
  author  = {Claude E. Shannon},
  title   = {Coding Theorems for a Discrete Source with a Fidelity Criterion},
  journal = {IRE National Convention Record},
  volume  = {7},
  number  = {4},
  pages   = {142--163},
  year    = {1959}
}

\clearpage
\appendix
\section{Supporting Measurements}
\label{app:detail}
All measurements in this appendix use the toy task. Values are the mean $\pm$
standard error over 8 seeds unless noted.

\begin{table}[H]
\centering\small
\caption{Scattered errors against bursts at the same average BER. The three burst
results are for bursts of 64\,b, 1\,kb and 16\,kb.}
\begin{tabular}{@{}lll@{}}
\toprule
Exposed bits & Scattered & Bursty \\
\midrule
All 8 element bits, $10^{-2}$ & $+4.6$\pct & $+6.1$ to $+7.0$\pct \\
Block scales, $10^{-3}$ & 8/8 diverge & 7/8, 4/8, 0/8 \\
bf16, all bits, $10^{-5}$ & 8/8 diverge & 8/8, 3/8, 0/8 \\
\bottomrule
\end{tabular}
\end{table}

\begin{table}[H]
\centering\small
\caption{Penalty (\%) against the share of rounds each site misses, over 80
rounds.}
\begin{tabular}{@{}lrr@{}}
\toprule
Rounds missed & Short outages & Long outages \\
\midrule
2\pct & $+0.8\pm1.3$ & $+0.4\pm1.0$ \\
10\pct & $+2.4\pm1.1$ & $+3.8\pm1.7$ \\
21 to 25\pct & $+6.7\pm1.7$ & $+15.9\pm3.9$ \\
\bottomrule
\end{tabular}
\end{table}

\begin{table}[H]
\centering\small
\caption{Penalty (\%) against training length. In the tracked rows the hub applies
the update as the sites decode it (4 seeds).}
\begin{tabular}{@{}lrr@{}}
\toprule
Link & 40 rounds & 640 rounds \\
\midrule
8-bit, unequal protection & $+0.8\pm0.4$ & $+28.6\pm3.7$ \\
\quad with resync every 10 rounds & $+0.1\pm0.5$ & $+0.9\pm1.5$ \\
4-bit, error-free & $+1.8\pm0.3$ & $+86.2\pm7.5$ \\
4-bit, error-free, tracked & & $-2.3\pm1.2$ \\
5-bit, error-free, tracked & & $-3.0\pm2.1$ \\
\bottomrule
\end{tabular}
\end{table}

\paragraph{Mantissa-only coding.} When the sign and exponent are delivered
exactly, one mantissa bit is enough on the toy task ($+0.9\pm0.4$\pct). Seven
mantissa bits at a 10\pct{} error rate cost $+0.4\pm0.2$\pct. The exponent is the
expensive part of this format. It takes 3.6 to 3.9 bits per value under ideal
entropy coding, where a shared block scale takes 0.25.

\end{document}